\RequirePackage{fix-cm}
\documentclass[smallextended]{svjour3}       
\smartqed  
\usepackage{graphicx}
\usepackage{subfigure}
 \usepackage{epstopdf}
 \usepackage{epsfig}
 \DeclareGraphicsExtensions{.pdf, .jpeg, .png}
\usepackage[colorlinks,linkcolor=red]{hyperref}   
\usepackage{bigstrut,multirow,rotating}      
\usepackage{booktabs}                             
\usepackage{enumerate}
\usepackage{amsmath}
\usepackage{bbm}
\usepackage{amsfonts,amssymb}
\usepackage{appendix}
\usepackage[numbers,sort&compress]{natbib}
\usepackage{url}
\usepackage[hyphenbreaks]{breakurl}

\begin{document}

\title{Computing large deviation prefactors of stochastic dynamical systems based on machine learning
}

\titlerunning{Computing large deviation prefactors}        

\author{Yang Li        \and
	    Shenglan Yuan$^{\ast}$        \and
        Linghongzhi Lu    \and
        Xianbin Liu      
}

\institute{Li Y. \at
               School of Automation, Nanjing University of Science and Technology, 200 Xiaolingwei Street, Nanjing 210094, China           
              \and
              Yuan S.$^{\ast}$ \at
              Center for Mathematical Sciences, Huazhong University of Science and Technology, 1037 Luoyu Road, Wuhan, 430074, China
              \and
              Lu L. and Liu X. \at
              State Key Laboratory of Mechanics and Control for Aerospace Structures, College of Aerospace Engineering, Nanjing University of Aeronautics and Astronautics, 29 Yudao Street, Nanjing 210016, China
              \\
              \\
              $^{\ast}$Corresponding author: shenglanyuan@hust.edu.cn
}

\date{Received: date / Accepted: date}

\maketitle

\begin{abstract}
In this paper, we present large deviation theory that characterizes the exponential estimate for rare events of stochastic dynamical systems in the limit of weak noise.  We aim to consider next-to-leading-order approximation for more accurate calculation of mean exit time via computing large deviation prefactors with the research efforts of machine learning. More specifically, we design a neural network framework to compute quasipotential, most probable paths and prefactors based on the orthogonal decomposition of vector field.  We corroborate the higher effectiveness and accuracy of our algorithm  with a practical example.
Numerical experiments demonstrate its powerful function in exploring internal mechanism of rare events triggered by weak random fluctuations.

\end{abstract}

\section{Introduction}
\label{intro}
The phenomena of rare events exit from the domain of attraction of a stable state induced by noise have been attracting increasing attention in recent years, ranging from physics \cite{MX,ZY,SB}, chemistry \cite{DM}, biology \cite{YLZ,YZD}, to engineering \cite{ZW,ZJ}. Even for weak noise, the rare events will occur almost surely, if the observations are performed on a sufficiently long time scale. The expected time required for observing this phenomena to occur, i.e., mean exit time, typically grows exponentially as the intensity of the random perturbations tends to zero.

Freidlin and Wentzell established large deviation theory to understand and analyze such dynamics in the limit of weak noise \cite{FW}. They proposed a vital concept of action functional to estimate the probability of the stochastic trajectory passing through the neighborhood of a given curve. The global minimum of the action functional is called quasipotential which exponentially dominates the magnitude of stationary probability distribution and mean exit time. However, the exponential estimate of large deviation theory is too rough since it smears all the possible polynomial coefficients prior to the exponent. Thus a reasonable estimation or an symptotic approximation of the large deviation prefactor is required, in order to gain a more accurate mean exit time.

Traditionally, shooting method is a feasible technique to compute the prefactor of mean exit time. Its idea is to derive a group of ordinary differential equations via WKB approximation and method of characteristics \cite{NK,MST,MS,R,MS97,BR22,BM}. Then the WKB prefactor and further the prefactor of mean exit time can be computed by integrating this group of equations. However, this method requires searching for the most probable path connecting the fixed point to the point with minimal quasipotential on the boundary, which is not an easy task, especially for high-dimensional systems.

We have witnessed in recent years a rapid development of data science and computer technology \cite{EW}. In view of the powerful nonlinear representation ability of machine learning, many researchers applied it to the investigation of stochastic dynamics. For example, machine learning methods can be used to discover stochastic dynamical systems from sample path data via nonlocal Kramers-Moyal formulas \cite{LD,LD22}, physics-informed neural networks \cite{KK,RV}, and variational inference \cite{O}. They are also used to solve physical quantities of stochastic dynamical systems via computing most probable path \cite{LDL,W}, quasipotential \cite{LXD,LLR,LYX}, and probability density \cite{XZL}. These experimental results show the potential applications of the combination between machine learning and stochastic dynamics.

In this research, our goal is to develop machine learning method to compute the large deviation prefactors in the highly complex nonlinear nondeterministic systems. We generate a specific
algorithm for training multilayer artificial networks and  perform complex computations through a learning process. The information-processing architecture is composed of neurons and synaptic weights converting electrical signals. It provides
a means of minimizing the error function, and even suggests a possible treatment of mean exit time. 

The structure of this article is arranged as follows. In Section \ref{LDP}, we simply introduce the large deviation theory and the quasipotential concept. In Section \ref{MET}, we describe the results for the prefactors of mean exit time in the cases of non-characteristic and characteristic boundaries. In Section \ref{Method},, we design the machine learning method for computing the prefactors. Numerical experiments are performed in Section \ref{nuex} to verify the effectiveness and accuracy of the algorithm.  Section \ref{conclu} presents the conclusion and innovations.

\section{Large deviation theory}\label{LDP}
We consider here $n$-dimensional stochastic differential equation (SDE) modeled by
\begin{equation}\label{SDE}
	dx=b(x)dt+\sqrt{\varepsilon}dB_t, \quad x\in\mathbb{R}^{n},
\end{equation}
where $b(x)=(b^1(x),b^2(x),\cdot\cdot\cdot,b^n(x))$ is the
vector field, $B_t=(B^1_t,B^2_t,\cdot\cdot\cdot,B^n_t)$ denotes Brownian motion, and $\varepsilon$ indicates a small noise intensity. Even though noise is weak, rare events, such as exit or transition problems induced by noise, will occur with probability one in sufficiently long time scale. Freidlin and Wentzell proposed large deviation theory to analyze and compute these phenomena \cite{FW}.

For the solution  $x(t), t\in[0,T]$ of system \eqref{SDE}, the Freidlin-Wentzell action functional is in the form of
\begin{equation*}
	\mathcal{S}[x]=\int_{0}^{T}\mathcal{L}(x,\dot{x})dt
\end{equation*}
with the corresponding Lagrangian $\mathcal{L}(x,\dot{x})=\frac{1}{2}|\dot{x}-b(x)|^{2}$.
The definition of quasipotential is given by
\begin{equation}\label{Vx}
	V(x):=\inf\limits_{0,T}\inf\limits_{\varphi\in C[0,T]}\big\{\mathcal{S}[\varphi]: \varphi(0)=\bar{x}, \varphi(T)=x\big\}.
\end{equation}
Note that the quasipotential of the fixed point equals zero, i.e., $V(\bar{x})=0$. The quasipotential $V(x)$ actually illustrates the possibility of the system appearing around $x$.

The associated Fokker-Planck equation of SDE \eqref{SDE} is formulated as
\begin{equation*}
	\partial_{t}p(x,t)=-\sum_{i=1}^{n}\partial_{i}(b^{i}(x)p(x,t))+\frac{1}{2}\varepsilon\sum_{i,j=1}^{n}\partial_{i}\partial_{j}p(x,t),
\end{equation*}
where $p(x,t)$ is the probability density. The stationary probability density $p_s(x)$ satisfies
\begin{equation}\label{spd}
	0=-\sum_{i=1}^{n}\partial_{i}(b^{i}(x)p_s(x))+\frac{1}{2}\varepsilon\sum_{i,j=1}^{n}\partial_{i}\partial_{j}p_s(x)=-\langle b(x),\nabla p_s(x)\rangle+\frac{1}{2}\varepsilon\langle\nabla p_s(x),\nabla p_s(x)\rangle.
\end{equation}
Assume that the stationary distribution is approached by the WKB approximation
\begin{equation}\label{WKB}
	p_{s}(x)\sim C(x)\exp\{-\varepsilon^{-1}V(x)\},
\end{equation}
where $C(x)$ stands for the WKB prefactor. Substituting the WKB approximation \eqref{WKB} into the stationary Fokker-Planck equation \eqref{spd} yields
Hamilton-Jacobi equation
\begin{equation}\label{HJE}
	H(x,\nabla V(x))=\langle b(x),\nabla V(x)\rangle+\frac{1}{2}\langle \nabla V(x),\nabla V(x)\rangle=0.
\end{equation}
Introducing the canonical momentum
\begin{equation*}
	p:=\frac{\partial\mathcal{L}(x,\dot{x})}{\partial\dot{x}}=\dot{x}-b(x),
\end{equation*}
the Hamiltonian $H(x,p)= \langle b(x),p\rangle +\frac{1}{2} \langle p,p\rangle$ is actually the Legendre transformation of the Lagrangian.

\begin{figure}\
	\centering
	\includegraphics[width=7cm]{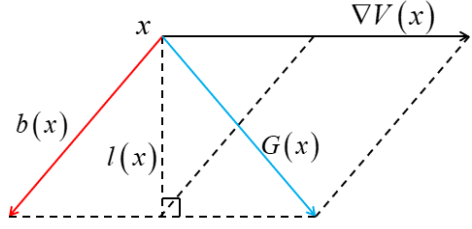}
	\caption{The orthogonal decomposition of the vector field $b(x)$.}
	\label{fig1}
\end{figure}

According to Eq. (\ref{HJE}), it follows that
\begin{equation*}
	\langle b(x)+\frac{1}{2}\nabla V(x),\nabla V(x)\rangle=0,
\end{equation*}
which thus infers the orthogonality relation
\begin{equation}\label{orthrela}
	b(x)+\frac{1}{2}\nabla V(x)\perp \nabla V(x).
\end{equation}
Define $l(x):=b(x)+\frac{1}{2}\nabla V(x)$.
It is immediately obvious that $b(x)=-\frac{1}{2}\nabla V(x)+l(x)$, which is concerned with an orthogonal decomposition of the vector field $b(x)$
as seen from Fig. \ref{fig1}. The vector field $G(x):=\frac{1}{2}\nabla V(x)+l(x)$ dominates the direction of the most probable path.

It should be pointed out that the WKB prefactor is determined by
\begin{equation*}
	C(x)\sim\sqrt{\frac{\det \nabla^{2}V(\bar{x})}{(2\pi\varepsilon)^{n}}}\exp\Big\{-\int_{-\infty}^{0}\text{div}l(\varphi_{t}^{x})dt\Big\},
\end{equation*}
where $\varphi^{x}$ represents the most probable exit path, ${\rm div}$ denotes the divergence, and $\bar{x}$ is the point at which $V$ attains its global minimum, i.e., the stable fixed point \cite[Section 3]{BR16}.

\section{Mean exit time}\label{MET}
Given an open subset domain $D\subset\mathbb{R}^{n}$, the first exit time is defined by the stopping time
\begin{equation*}
	\tau_{D}^{\varepsilon}:=\inf\{t\geq0: x_{t}\notin D, x(0)=\bar{x}\},\quad \text{for}\,\, \bar{x}\in D.
\end{equation*}
Estimating the mean first exit time $\mathbb{E}\tau_{D}^{\varepsilon}$ starting from the stable fixed point $\bar{x}$ requires minimizing the action functional $\mathcal{S}[x]$ over all exit times in terms of the Freidlin-Wentzell large deviation theory. Performing this minimization over paths from the point $\bar{x}$ to the basin boundary $\partial D$ yields the Arrhenius law
\begin{equation*}
	\lim_{\varepsilon\rightarrow0}\varepsilon\ln\mathbb{E}\tau_{D}^{\varepsilon}=\inf_{x\in\partial D}V(x).
\end{equation*}
Therefore,
\begin{equation}\label{metLV}
	\mathbb{E}\tau_{D}^{\varepsilon}=L_{D}^{\varepsilon}\exp\{\varepsilon^{-1}\inf_{x\in\partial D}V(x)\}.
\end{equation}
We compute the above prefactor $L_{D}^{\varepsilon}$ in two different cases.

\textbf{Case A.} non-characteristic boundary \cite[Section 4.1]{BR22}.

Assume that the domain $D$ is an open, smooth and connected subset of $\mathbb{R}^{n}$ satisfying the following conditions, where $n(y)$ indicates the exterior normal vector at $y\in\partial D$.
\begin{description}
	\item[(A1)] The deterministic system $\dot{x}=b(x)$ possesses a unique fixed point $\bar{x}$ in $D$, which attracts all the trajectories started from $D$, and $\langle b(y),n(y)\rangle<0$ for all $y\in \partial D$.
	\item[(A2)] The function $V$ is $C^1$-continuous in $D$; for any $x\in\bar{D}$, the most probable path $\varphi_{t}^{x}$ goes to $\bar{x}$ as $t\rightarrow-\infty$; and $\langle \nabla V(y),n(y)\rangle >0$ for all $y\in \partial D$.
	\item[(A3)]  The minimum of $V$ over $\partial D$ is reached at a single point $x^{*}$, at which
	\begin{equation*}
		\mu^{*}=\langle \frac{1}{2} \nabla V(x^{*})+l(x^{*}),n(x^{*})\rangle>0,
	\end{equation*}
	and the quadratic form $h^{*}: \xi\mapsto\langle\xi,\nabla^{2}V(x^{*})\xi\rangle$ has positive eigenvalues on the hyperplane $n(x^{*})^{\perp}=\{\xi\in\mathbb{R}^{n}: \langle\xi,n(x^{*})\rangle=0\}$.
\end{description}

If $\langle b(y),n(y)\rangle<0$ for all $y\in \partial D$, the boundary is said to be non-characteristic. This condition ensures
that the dynamical trajectories  started from the closure $\bar{D}$ will remain $D$ and that the vector field is transverse with the boundary. By Assumption (A1), we derive the integral formula \cite{BR22,BR16}
\begin{equation*}
	\lambda_{D}^{\varepsilon}=\int\limits_{x\in \partial D}\langle \frac{1}{2} \nabla V(x)+l(x),n(x)\rangle C(x)\exp\{-\varepsilon^{-1}V(x)\}dx
\end{equation*}
for the exit rate $\lambda_{D}^{\varepsilon} =\mathbb{E} [\tau_{D}^{\varepsilon}] ^{-1}$. Using the second-order expansion of $V$ in the neighborhood of $x^{*}$ in this formula, we obtain the equivalent relation of the prefactor
\begin{equation}\label{LcaseA}
	\begin{aligned}
		L_{D}^{\varepsilon}&\sim\frac{1}{C(x^{\ast})\mu^{\ast}}\sqrt{\frac{\det h^{\ast}}{(2\pi\varepsilon)^{n-1}}}  \\
		&\sim\frac{1}{\mu^{\ast}}\sqrt{\frac{2\pi\varepsilon\det h^{\ast}}{\det\nabla^{2}V(\bar{x})}}\exp\Big\{-\int_{-\infty}^{0}\text{div}l(\varphi_{t}^{x^{\ast}})dt\Big\},
	\end{aligned}
\end{equation}
where we use
\begin{equation*}
	C(x^*)\sim\sqrt{\frac{\det \nabla^{2}V(\bar{x})}{(2\pi\varepsilon)^{n}}}\exp\Big\{-\int_{-\infty}^{0}\text{div}l(\varphi_{t}^{x^*})dt\Big\}.
\end{equation*}

\textbf{Case B.} characteristic boundary \cite[Section 4.2]{BR22}.

The basin domain $D$ is characteristic in the sense that $\langle b(y),n(y)\rangle=0$ for all $y\in D$. We consider the metastable case that the deterministic system $\dot{x}=b(x)$ possesses two stable fixed points $\bar{x}_1$ and $\bar{x}_2$, whose basins of attractions are separated by a smooth hypersurface $S$. We consider the exit events from the basin of attraction $D$ of $\bar{x}_1$ and formulate the following set of assumptions.
\begin{description}
	\item[(B1)] All the trajectories of the deterministic system $\dot{x}=b(x)$ started on $S$ remain in $S$ and
	converge to a single fixed point $x^{\ast}\in S$; in addition, the Jacobi matrix $\nabla b(x^{\ast})$ possesses $n-1$ eigenvalues with negative real part and a single positive eigenvalue $\lambda^{*}$.
	\item[(B2)] Indicating the quasipotential with respect to $\bar{x}_1$ as $V$, there exists a unique (up to time shift) trajectory $\rho=(\rho_t)_{t\in\mathbb{R}}\subset D$ such that
	\begin{equation*}
		\lim_{t\rightarrow-\infty}\rho_t=\bar{x}_{1},\quad \lim_{t\rightarrow+\infty}\rho_t=x^{*},\quad\text{and}\quad V(x^{*})=\mathcal{S}_{-\infty,+\infty}[\rho].
	\end{equation*}
	\item[(B3)] $V$ is smooth in the neighborhood of $\rho=(\rho_t)_{t\in\mathbb{R}}$, and the vector field $l$ defined by  $b(x)=-\frac{1}{2}\nabla V(x)+l(x)$ satisfies the orthogonality relation.
\end{description}

In this context, the quasipotential $V$ achieves its minimum on $S$ at the point $x^{*}$. Moreover, the path $\rho$ is called the most probable exit path and it satisfies
\begin{equation*}
	\forall t\in\mathbb{R}, \ \dot{\rho}_t=\frac{1}{2}\nabla V(\rho_t)+l(\rho_t).
\end{equation*}
For any $t\in\mathbb{R}$, it coincides with the path $(\varphi_s^{x})_{s\leq0}$ connecting $\bar{x}$ with $x=\rho_t$, in the sense that
\begin{equation*}
	\forall s\leq0,\quad\varphi_{s}^{x}=\rho_{s+t}.
\end{equation*}
In order to describe the prefactor $L_D^{\epsilon}$ in this case, we formulate the following supplementary assumption.
\begin{description}
	\item[(B4)] The matrix $H^{\ast}=\lim_{t\rightarrow+\infty}\nabla^{2}V(\rho_t)$ exists and has $n-1$ positive eigenvalues and 1
	negative eigenvalue.
\end{description}

Based on these four assumptions, an asymptotic formula for estimating the expected time taken by the process to exit $D$ was obtained in \cite[Eq. (1.10)]{BR16}, i.e.,
\begin{equation}\label{LcaseB}
	L_{D}^{\varepsilon}\sim\frac{\pi}{\lambda^{\ast}}\sqrt{\frac{|\det H^{\ast}|}{\det\nabla^{2}V(\bar{x})}}\exp\Big\{-\int_{-\infty}^{\infty}\text{div}l(\varphi_{t})dt\Big\}.
\end{equation}

\section{Method}
\label{Method}
\begin{figure}
	\centering
	\includegraphics[width=7cm]{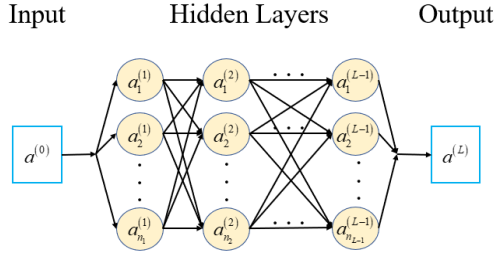}
	\caption{Architecture of the neural network with $L-1$ hidden layers. The input is $a^{(0)}$, the output is $a^{(L)}$, and $a^{(l)}_{j}$ denotes the value of the $j$-th neuron in the $l$-th hidden layer for $j=1, \cdot\cdot\cdot, n_l$, $l=1, \cdot\cdot\cdot, L-1$.}
	\label{fig2}
\end{figure}

To compute the large deviation prefactors defined in Section \ref{MET}, we are devoted to developing a machine learning method inspired by Ref. \cite{LLR}. The critical idea is to design the graphic of the feedforward multilayer neural network shown in Fig. \ref{fig2} to realize the orthogonal decomposition of the vector field. The neural network is made up of three basic components: an input vector $a^{(0)}=x\in\mathbb{R}^{n}$, a set of hidden layers with some synapses connecting neurons, and an output layer $a^{(L)}=(\widehat{V}_{\theta},l_{\theta})\in\mathbb{R}^{n+1}$.
For $l=1, \cdot\cdot\cdot, L-1$, there are $n_l$ neurons in the $l$-th hidden layer. The subscript $\theta$ denotes the training parameters in the neural network.

Note that the quasipotential is similar to a quadratic function around the stable fixed point $\bar{x}$. In order to guarantee that the quasipotential $V_{\theta}$ and its gradient $|V_{\theta}|$ are all unbounded, we define the quasipotential $V_{\theta}= \widehat{V}_{\theta} +|x-\bar{x}|^{2}$. To train the neural network, we construct the loss function $L:=L_{\text{dyn}}+\gamma_{1}L_{\text{orth}}+\gamma_{2}L_{0}$, where $\gamma_{1}$ and $\gamma_{2}$ are weighted parameters harmonizing the proportion of several parts of loss function.

In order to design the loss function, we randomly select points $\{x_i, i=1,2,\cdot\cdot\cdot,N\}$ in the domain of interest in $n$-dimensional Euclidean space $\mathbb{R}^{n}$. Since the vector field can be decomposed as $b(x)=-\frac{1}{2}\nabla V(x)+l(x)$, the first part of the loss function can be set as
\begin{equation*}
	L_{\text{dyn}}=\frac{1}{N}\sum_{i=1}^{N}[b(x_{i})+\frac{1}{2}\nabla V_{\theta}(x_i)-l_\theta(x_i)]^2.
\end{equation*}
Due to the orthogonal condition $\nabla V(x)\perp l(x)$, we specify the second part of the loss function as
\begin{equation*}
	L_{\text{orth}}=\frac{1}{N}\sum_{i=1}^{N}\frac{[\nabla V_{\theta}(x_i)\cdot l_{\theta}(x_i)]^2}{|\nabla V_{\theta}(x_i)|^2\cdot|l_{\theta}(x_i)|^2+\delta},
\end{equation*}
where we add  a sufficiently small parameter $\delta>0$, avoiding the divergence of $L_{\text{orth}}$ in the case of $l_{\theta}(x)=0$.
The quasipotential is zero at the fixed point, so we fix $L_0=V_{\theta}(\bar{x})^{2}$.

To examine the accuracy of the algorithm, we define the approximation error functions for the quasipotential and the rotational component of vector field by
\begin{equation*}
	e_{V}=\frac{\max_x|V_{\theta}(x)-V(x)|^2}{\max_x|V(x)|^2}\quad\text{and}\quad e_l=\frac{\max_x|l_{\theta}(x)-l(x)|^2}{\max_x|l(x)|^2},
\end{equation*}
where $V_{\theta}$ and $l_{\theta}$ are the prediction results of the neural network, and $V$ and $l$  mean the true results. According to
the expression of $L_{D}^{\varepsilon}$ for the mean first exit time in Section \ref{MET}\quad, the computation of the prefactor essentially depends on the results of the most probable path $\varphi_{t}$ and the rotational component $l(x)$. After training the neural network, the most probable path satisfies
\begin{equation*}
	\dot{\varphi}_{t}=b(\varphi_{t})+\nabla V_{\theta}(\varphi_{t}).
\end{equation*}
For numerical convenience, we transform the parameter of this equation into the arc length $\sigma$ of the most probable path. Then
\begin{equation*}
	\frac{d\sigma}{dt}=|\dot{\varphi}|=|b(\varphi)+\nabla V_{\theta}(\varphi)|=|b(\varphi)|,
\end{equation*}
where the last equality comes from $|G(x)|=|b(x)|$ as in Fig. \ref{fig1}. Therefore, the most probable path satisfies
\begin{equation*}
	\frac{d\varphi}{d\sigma}=\frac{b(\varphi)+\nabla V_{\theta}(\varphi)}{|b(\varphi)|}.
\end{equation*}
We can obtain the most probable path from the inverse time integration of the above equation starting from the endpoint $x^{*}$ until reaching the neighborhood of the fixed point.

With regard to \textbf{Case A}, we choose the point $x^{*}$ at which the quasipotential reaches its minimum value on the boundary $\partial D$. We calculate $\nabla l_{\theta}(x)$ in virtue of the neural network results of $l_{\theta}(x)$. Then the integral in $L_{D}^{\varepsilon}$  can be transformed into
\begin{equation*}
	\int_{-\infty}^{0}\text{div}l_{\theta}(\varphi_{t}^{x^{*}})dt=\int_{0}^{L}\frac{\text{div}l_{\theta}(\varphi_{\sigma}^{x^{*}})}{|b(\varphi_{\sigma}^{x^{*}})|}d\sigma.
\end{equation*}
The calculation of the WKB prefactor $C(x^{*})$ is already included in the calculation of $L_{D}^{\varepsilon}$ for \textbf{Case A}. It will no longer be considered separately.

In \textbf{Case B}, we take $x^{*}$ as the saddle point $x_{\text{sad}}$. For the sake of efficient numerical implementation,
our actual operation is to take $x^{*}=x_{\text{sad}}+\delta_{1}e_{u}$ with the help of a small parameter $\delta_{1}>0$, where $e_{u}$ is
the unstable direction of $x_{\text{sad}}$. For a small parameter $\delta_{2}>0$, the inverse time integration terminates
when it enters into the $\delta_{2}$-neighborhood of the fixed point $\bar{x}$.  For \textbf{Case B}, we rewrite the integral in $L_{D}^{\varepsilon}$ into
\begin{equation*}
	\int_{-\infty}^{\infty}\text{div}l_{\theta}(\varphi_{t})dt=\int_{0}^{L}\frac{\text{div}l_{\theta}(\varphi_{\sigma})}{|b(\varphi_{\sigma})|}d\sigma.
\end{equation*}

\section{Numerical experiments}\label{nuex}
\begin{figure}
	\centering
	\includegraphics[width=7cm]{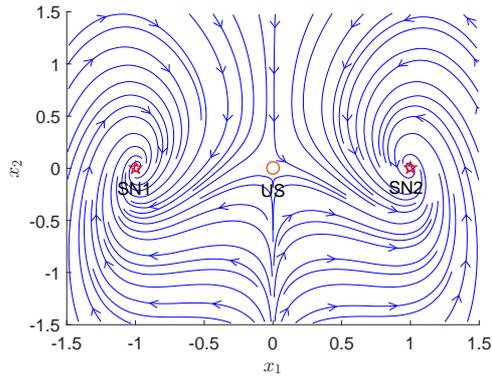}
	\caption{The flow of the vector field $b(x)$ as in \eqref{bx}. $\text{SN1}:=(-1,0)$ and $\text{SN2}:=(1,0)$ denote two stable fixed points and $\text{US}:=(0,0)$ indicates a saddle.}
	\label{fig3}
\end{figure}

To verify the effectiveness and accuracy of the algorithm, we investigate an example with an explicit quasipotential
\begin{equation*}
	V(x_1,x_2)=v_{1}(x_1)+v_{2}(x_2),
\end{equation*}
where $v_{1}(x_1)=\frac{1}{2}x_1^{4}-x_1^{2}$ and $v_{2}(x_2)=\alpha x_2^{2}$.
Define the vector field $b(x)=-\frac{1}{2}\nabla V(x)+l(x)$ with
\begin{equation*}
	l(x_1,x_2)=c(x_1,x_2)\left(
	\begin{array}{c}
		-v_{2}'(x_2)\\
		v_{1}'(x_1)\\
	\end{array}
	\right),
\end{equation*}
where $c(x_1,x_2)=\beta x_1$. It follows from
\begin{equation*}
	\langle\nabla V(x), l(x)\rangle=c(x_1,x_2)\left(
	\begin{array}{cc}
		v_{1}'(x_1) & v_{2}'(x_2) \\
	\end{array}
	\right)\left(
	\begin{array}{c}
		-v_{2}'(x_2)\\
		v_{1}'(x_1)\\
	\end{array}
	\right)=c(x_1,x_2)\big(v_{1}'(x_1)v_{2}'(x_2)-v_{1}'(x_1)v_{2}'(x_2)\big)=0
\end{equation*}
that $\nabla V(x)\perp l(x)$. When $\alpha=0.5$ and $\beta=3$,
\begin{equation}\label{bx}
	b(x)=\left(
	\begin{array}{c}
		x_1-x_1^{3}-3x_1x_2 \\
		6x_1^{4}-6x_1^{2}-\frac{1}{2}x_2 \\
	\end{array}
	\right).
\end{equation}
If $b(x)=0$, then we know that there are three fixed points $(0,0)$ and $(\pm1,0)$.
To classify those fixed points, we take into account the Jacobian matrix
\begin{equation*}
	J=\left(
	\begin{array}{cc}
		1-3x_{1}^{2}-3x_{2} & -3x_{1} \\
		24x_{1}^{3}-12x_{1} & -\frac{1}{2} \\
	\end{array}
	\right).
\end{equation*}
The Jacobian matrix at $(\pm1,0)$ is given by
\begin{equation*}
	J_{(\pm1,0)}=\left(
	\begin{array}{cc}
		-2 & \mp3 \\
		\pm12 & -\frac{1}{2} \\
	\end{array}
	\right).
\end{equation*}
The characteristic equation is $\lambda^{2}+\frac{5}{2}\lambda+37=0$, which has complex roots
\begin{equation*}
	\lambda_{\pm}=\frac{-5\pm9\sqrt{7}i}{4}.
\end{equation*}
Thus the two fixed points $(\pm1,0)$ are asymptotically stable. Using the Jacobian matrix
\begin{equation*}
	J_{(0,0)}=\left(
	\begin{array}{cc}
		1 & 0 \\
		0 & -\frac{1}{2} \\
	\end{array}
	\right),
\end{equation*}
the eigenvalues for $(0,0)$ are $\lambda_{1}=1$ and $\lambda_{2}=-\frac{1}{2}$. Since one eigenvalue is positive and the other is  negative, the origin is an unstable saddle point. We sketch a phase portrait for the deterministic system $\dot{x}=b(x)$ in Fig. \ref{fig3}. The eigenvector for $\lambda_{2}=-\frac{1}{2}$ is $(0,1)^{T}$. The stable manifold  is the $x_2$-axis, which forms the boundary of the left and right basins of attraction. Thanks to the symmetry of the image, we need only consider the exit problem starting from the left side.

The selection of superparameters in algorithm consists of 6 hidden layers with 20 neurons per layer. We choose the Adam optimizer with the learning rate 0.002.  The nonlinear activation function for hidden layer is the hyperbolic tangent function $\tanh$. The activation function for output is identity function. The number of training epochs is 100000. We fix $\gamma_{1}=1$, $\gamma_{2}=0.1$, $\delta=0.001$, $\delta_1=0.05$ and $\delta_2=0.01$. We randomly select $N=100, 1000, 10000, 100000$ points on the region $[-1.5,0]\times[-0.8,0.8]$. After completing the training process, we calculate the approximation errors ${e}_{V}$ and ${e}_{l}$ with the results shown in Table \ref{table1}. In the beginning, the errors ${e}_{V}$ and ${e}_{l}$ gradually decrease as the amount of data increases from $N=100$ to $N=10000$. However, it can be observed that the change is not significant for $N=100000$. At this point, the accuracy of the algorithm is limited by other parameters.

\begin{table}[htbp]
	\centering
	\caption{Approximation errors}
	\begin{tabular}{ccccc}
		\toprule
		\specialrule{0em}{1pt}{1pt}
		$N$      & 100   & 1000   & 10000   & 100000     \\
		\midrule
		\specialrule{0em}{1pt}{1pt}
		${e}_{V}$      & 0.3122\% & 0.1392\% & 0.1267\% & 0.0681\%  \\
		\specialrule{0em}{1pt}{1pt}
		${e}_{l}$      & 1.3692\% & 0.2127\% & 0.0738\% & 0.0965\%  \\
		\bottomrule
	\end{tabular}
	\label{table1}
\end{table}

\begin{figure}
	\centering
	\subfigure[Learned quasipotential]
	{\includegraphics[width=7cm]{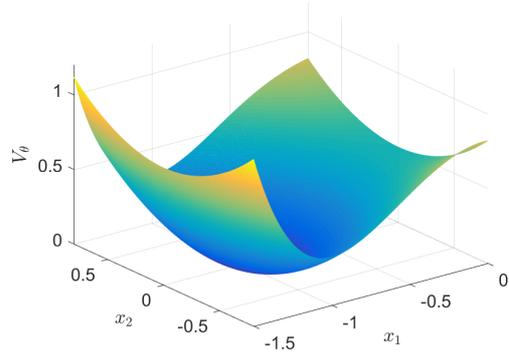}}
	\subfigure[True quasipotential]
	{\includegraphics[width=7cm]{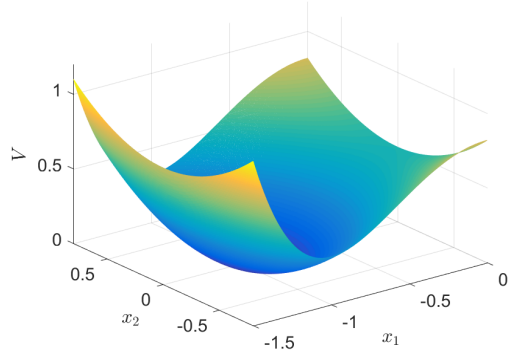}}
	\caption{Comparison between learned and true quasipotential.}
	\label{fig4}
\end{figure}

Comparing the numerical results of learned and true quasipotential displayed in Fig. \ref{fig4} for $N=100000$, it can be seen that the two pictures are consistent with each other.

When $N=100000$, we compute the prefactors for \textbf{Case A} and \textbf{Case B}.

\textbf{Case A.} non-characteristic boundary.

\begin{figure}
	\centering
	\includegraphics[width=7cm]{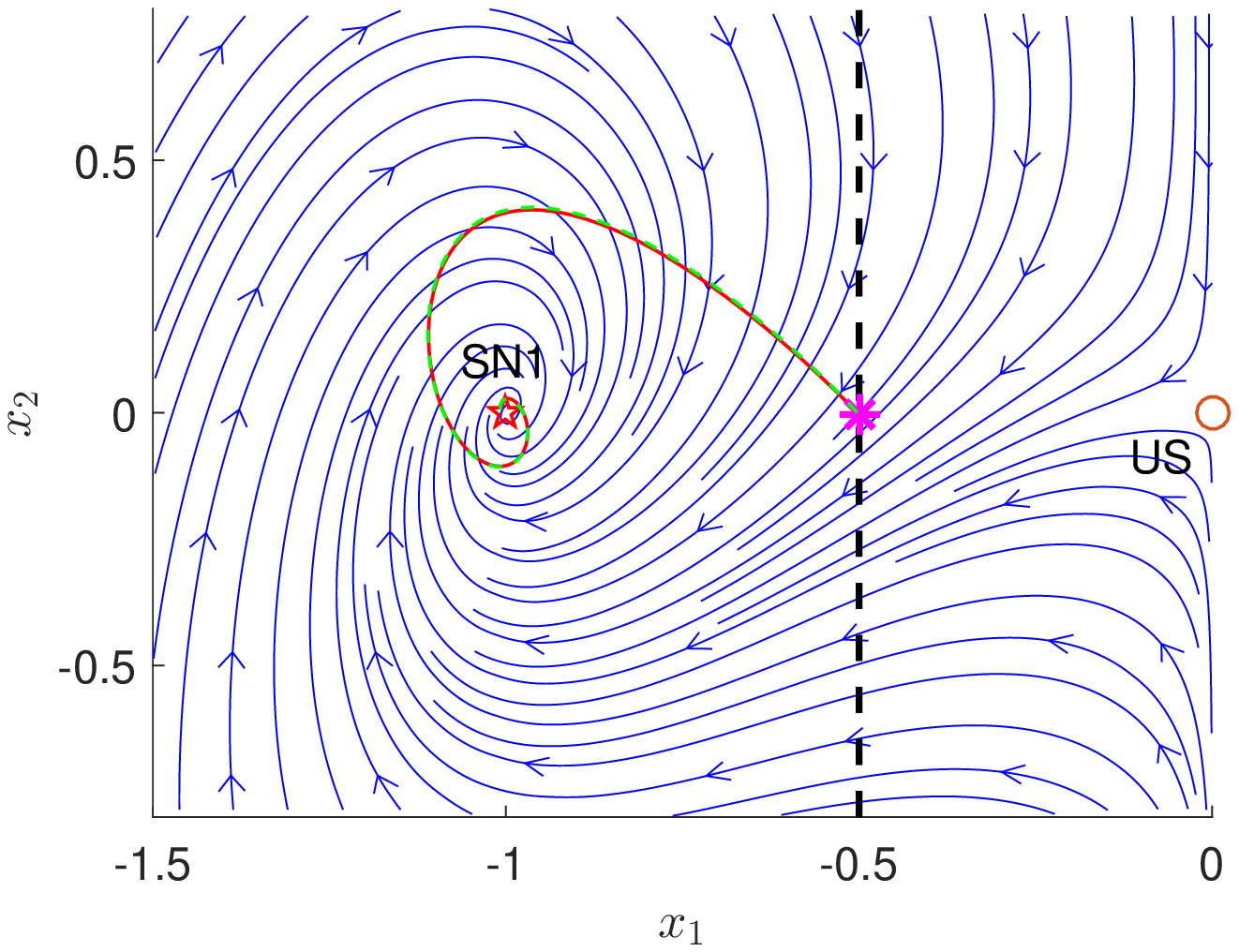}
	\caption{Comparison between the most probable exit paths for \textbf{Case A} computed by machine learning and true system, denoted by red and dashed green curves, respectively. The dashed black line indicates the boundary $x_1=-0.5$. The pink star $x^{*}=(-0.4990,-0.0048)$  is the exit point.}
	\label{fig5}
\end{figure}

As depicted in Fig. \ref{fig5}, we choose the dashed black line $x_1=-0.5$ as the non-characteristic boundary for the basin of attraction of SN1. The pink star $x^{*}=(-0.4990,-0.0048)$ denotes the exit point, which represents the minimum of the quasipotential on the boundary. By utilizing the results of neural network and true system starting from $x^{*}$, we simulate the most probable paths marked as red and dashed green curves, respectively. As indicated in Fig. \ref{fig5}, the two curves match very well.

Define $\bar{H}:=\nabla^{2}V(\bar{x})$, which satisfies the algebraic Riccati equation
\begin{equation*}
	2\bar{H}^{2}=\bar{Q}^{T}\bar{H}+\bar{H}\bar{Q},
\end{equation*}
where the matrix
\begin{equation*}
	\bar{Q}=-\nabla b(\bar{x})=\left(
	\begin{array}{cc}
		2 & -\alpha\beta \\
		2\beta & \alpha \\
	\end{array}
	\right).
\end{equation*}
Then we have
\begin{equation*}
	\bar{H}=\left(
	\begin{array}{cc}
		2 & 0 \\
		0 & \alpha \\
	\end{array}
	\right).
\end{equation*}
Owing to the calculation results of $l_{\theta}(x)$, we can get $L_{D}^{\varepsilon}\sim2.1525\sqrt{\varepsilon}$.
The true values of $l(x)$ can lead to $L_{D}^{\varepsilon} \sim2.2164 \sqrt{\varepsilon}$. The approximate values of $L_{D}^{\varepsilon}$  are very close in two different situations.

\begin{figure}
	\centering
	\includegraphics[width=7cm]{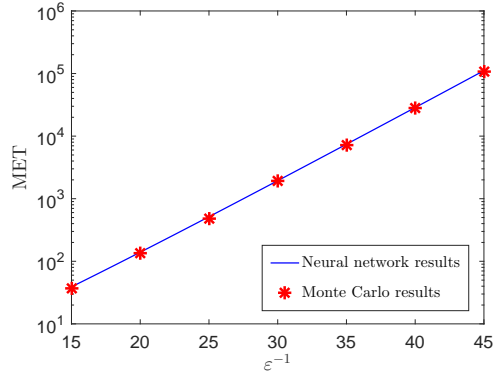}
	\caption{Comparison between mean exit times for \textbf{Case A} via machine learning and Monte Carlo simulation, denoted by blue curve and red star, respectively.}
	\label{fig6}
\end{figure}

Substituting the prefactor $L_{D}^{\varepsilon} \sim2.1525 \sqrt{\varepsilon}$ and the quasipotential $V_{\theta}(x^{*})=0.2835$ into the expression of mean exit time $\mathbb{E}\tau_{D}^{\varepsilon} =L_{D}^{\varepsilon}\exp\{\varepsilon^{-1}V_{\theta}(x^{*})\}$, we plot the compared graph between mean exit times via machine learning and Monte Carlo simulation in Fig. \ref{fig6}. The red asterisk denotes the results of Monte Carlo simulation, which is completely consistent with the learning results and has very high accuracy.

\textbf{Case B.} characteristic boundary.

\begin{figure}
	\centering
	\includegraphics[width=7cm]{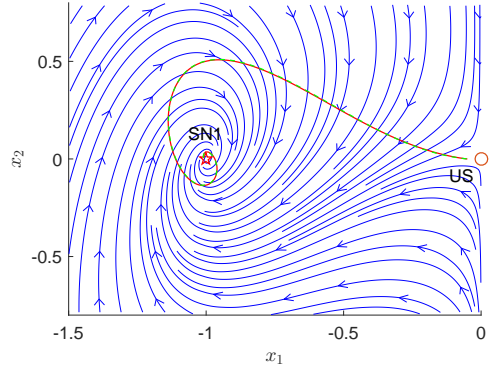}
	\caption{Comparison between the most probable exit paths for \textbf{Case B} computed by machine learning and true system, denoted by red and dashed green curves, respectively.}
	\label{fig7}
\end{figure}

We treat $x_2$-axis, i.e., the stable manifold of $\text{US}$, as the boundary of the left and right basins of attraction. On the grounds of neural network, we acquire the most probable path by the dashed green curve that coincides with the red curve of the true system, as shown in Fig. \ref{fig7}.

\begin{figure}
	\centering
	\includegraphics[width=7cm]{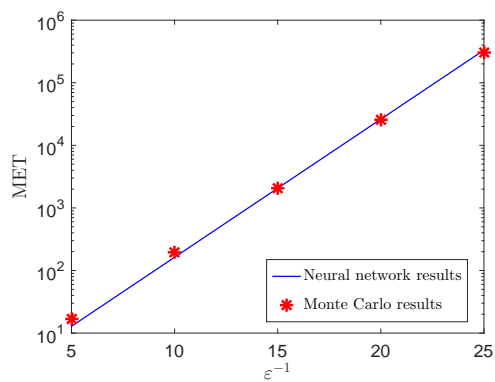}
	\caption{Comparison between mean exit times for \textbf{Case B} via machine learning and Monte Carlo simulation, denoted by blue curve and red star, respectively.}
	\label{fig8}
\end{figure}

Define $H^{*}:=\nabla^2 V(\text{US})$, which satisfies the algebraic Riccati equation
\begin{equation*}
	2{H^{*}}^{2}={Q^{*}}^{T}H^{*}+H^{*}Q^{*},
\end{equation*}
where the matrix
\begin{equation*}
	Q^{*}=-\nabla b(\text{US})=\left(
	\begin{array}{cc}
		-1 & 0 \\
		0 & \alpha \\
	\end{array}
	\right).
\end{equation*}
Then we have
\begin{equation*}
	H^{*}=\left(
	\begin{array}{cc}
		-1 & 0 \\
		0 & \alpha \\
	\end{array}
	\right).
\end{equation*}
We evaluate  $L_{D}^{\varepsilon}\sim1.0121$ by the use of the calculation result of $l_{\theta}(x)$, and obtain $L_{D}^{\varepsilon}\sim1.0958$ on the basis of the calculation result of $l(x)$.
Putting $V_{\theta}(\text{US})=0.5081$ and $L_{D}^{\varepsilon}\sim1.0121$ into
the mean exit time $\mathbb{E}\tau_{D}^{\varepsilon}=L_{D}^{\varepsilon}\exp\{\varepsilon^{-1}V(x^{*})\}$,
the comparison chart between the neural network and Monte Carlo results is presented in Fig. \ref{fig8}. It is worth notice that the training result of neural network is highly consistent with the experimental result of Monte Carlo.

\section{Conclusion and future perspective}
\label{conclu}
In this work, we developed a machine learning method to establish the large deviation prefactors of stochastic dynamical systems. In particular, we first introduced large deviation theory and the results of prefactors of mean exit time for non-characteristic and characteristic boundaries. Then we provided a new machine learning method to compute the prefactors based on the orthogonal decomposition of the vector field. The successful application of the algorithm to a toy model illustrated its effectiveness of tackling mean exit time accurately.

Our neural networks and algorithm can be effectively extended to and sophisticated  high-dimensional circumstances with computational complexity. They can also be generalized to deal with other quantities of rare events such as stationary probability distribution and escape probability. More importantly, our innovative approach may bring forth new ideas for improving the performance of searching for the most probable path by machine learning or data mining.

\section*{Acknowledgement}
This research was supported by Natural Science Foundation of Jiangsu Province (grant BK20220917) and National Natural Science Foundation of China (grant 12001213).

\section*{Data Availability Statement}
The data that support the findings of this study are openly available in GitHub \\ \url{https://github.com/liyangnuaa/Computing-large-deviation-prefactors}.

%
%


\end{document}